\documentclass[conference]{IEEEtran}
\IEEEoverridecommandlockouts

\usepackage{cite}
\usepackage{amsmath,amssymb,amsfonts}
\usepackage{graphicx}
\usepackage{textcomp}
\usepackage{xcolor}
\usepackage{hyperref}
\usepackage{soul}
\usepackage{algorithm}
\usepackage{algpseudocode}
\usepackage{graphicx}
\usepackage{multirow}
\usepackage{booktabs}
\usepackage{rotating}
\usepackage{tikz}

\def\BibTeX{{\rm B\kern-.05em{\sc i\kern-.025em b}\kern-.08em
    T\kern-.1667em\lower.7ex\hbox{E}\kern-.125emX}}

\usepackage{enumitem}
\setlist[itemize]{leftmargin=1em}

\newcommand*\circled[1]{\tikz[baseline=(char.base)]{             \node[shape=circle,fill,inner sep=1pt] (char) {\textcolor{white}{#1}};}}

\newcommand{\projname}{\textsc{Labeling Copilot}}

\begin{document}

\title{\textsc{Labeling Copilot}: A Deep Research Agent for Automated Data Curation in Computer Vision
\thanks{This work was supported in part by the NSF research grant \#2320952, \#2117439, \#2112606, and \#2117439.}
}

\pagestyle{plain} 

\author{
\IEEEauthorblockN{Debargha Ganguly\textsuperscript{*1,3}, Sumit Kumar\textsuperscript{*1,4}, Ishwar Balappanawar\textsuperscript{*1,4}, Weicong Chen\textsuperscript{*3}, Shashank Kambhatla\textsuperscript{1,5}}
\IEEEauthorblockA{Srinivasan Iyengar\textsuperscript{2}, Shivkumar Kalyanaraman\textsuperscript{2}, Ponnurangam Kumaraguru\textsuperscript{4}, Vipin Chaudhary\textsuperscript{3}}
\\
\IEEEauthorblockA{\textsuperscript{1}Microsoft Research \textsuperscript{2}Microsoft Corporation \textsuperscript{3}Case Western Reserve University\\
\textsuperscript{4}IIIT Hyderabad \textsuperscript{5}University of Pennsylvania\\
\{debargha, weicong, vipin\}@case.edu, \{sriyengar, shkalya\}@microsoft.com, sumit.k@research.iiit.ac.in,\\
ishwar.balappanawar@students.iiit.ac.in, pk.guru@iiit.ac.in, skamb@seas.upenn.edu}\\
}

\maketitle

\begingroup
  \renewcommand\thefootnote{\fnsymbol{footnote}}
  \setcounter{footnote}{1}
  \footnotetext{These authors contributed equally to this work.}
\endgroup

\begin{abstract}

Curating high-quality, domain-specific datasets is a major bottleneck for deploying robust vision systems, requiring complex trade-offs between data quality, diversity, and cost when researching vast, unlabeled data lakes. We introduce Labeling Copilot, the first data curation deep research agent for computer vision. A central orchestrator agent, powered by a large multimodal language model, uses multi-step reasoning to execute specialized tools across three core capabilities: (1) \textbf{Calibrated Discovery} sources relevant, in-distribution data from large repositories; (2) \textbf{Controllable Synthesis} generates novel data for rare scenarios with robust filtering; and (3) \textbf{Consensus Annotation} produces accurate labels by orchestrating multiple foundation models via a novel consensus mechanism incorporating non-maximum suppression and voting. Our large-scale validation proves the effectiveness of Labeling Copilot's components. The Consensus Annotation module excels at object discovery: on the dense COCO dataset, it averages 14.2 candidate proposals per image—nearly double the 7.4 ground-truth objects—achieving a final annotation mAP of 37.1\%. On the web-scale Open Images dataset, it navigated extreme class imbalance to discover 903 new bounding box categories, expanding its capability to over 1500 total. Concurrently, our Calibrated Discovery tool, tested at a 10-million sample scale, features an active learning strategy that is up to 40x more computationally efficient than alternatives with equivalent sample efficiency. These experiments validate that an agentic workflow with optimized, scalable tools provides a robust foundation for curating industrial-scale datasets.

\end{abstract}

\section{Introduction}
The remarkable progress in Computer Vision (CV) has been fundamentally enabled by large-scale, high-quality, domain-specific datasets such as ImageNet~\cite{deng2009imagenet}, COCO~\cite{lin2014microsoft}, and Open Images~\cite{kuznetsova2020open}. The curation of these datasets, however, remains a persistent bottleneck that limits the scalable deployment and real-world impact of CV models~\cite{paullada2021data, liang2022mind}. Automating this complex workflow using foundation models~\cite{xiao2024florence} and agentic systems have become an active area of research~\cite{gao2025survey, fang2025comprehensive}. While popular models like GroundingDINO~\cite{liu2023grounding} and SAM~\cite{kirillov2023segment} excel on benchmarks, they are yet to scale to the knowledge-intensive task of curating production-grade datasets. This intricate process is not a linear pipeline but a dynamic challenge requiring intelligent decision-making, making it an ideal application for an agentic framework~\cite{sumers2023cognitive}.

\begin{figure*}
    \centering
    \includegraphics[width=1\linewidth]{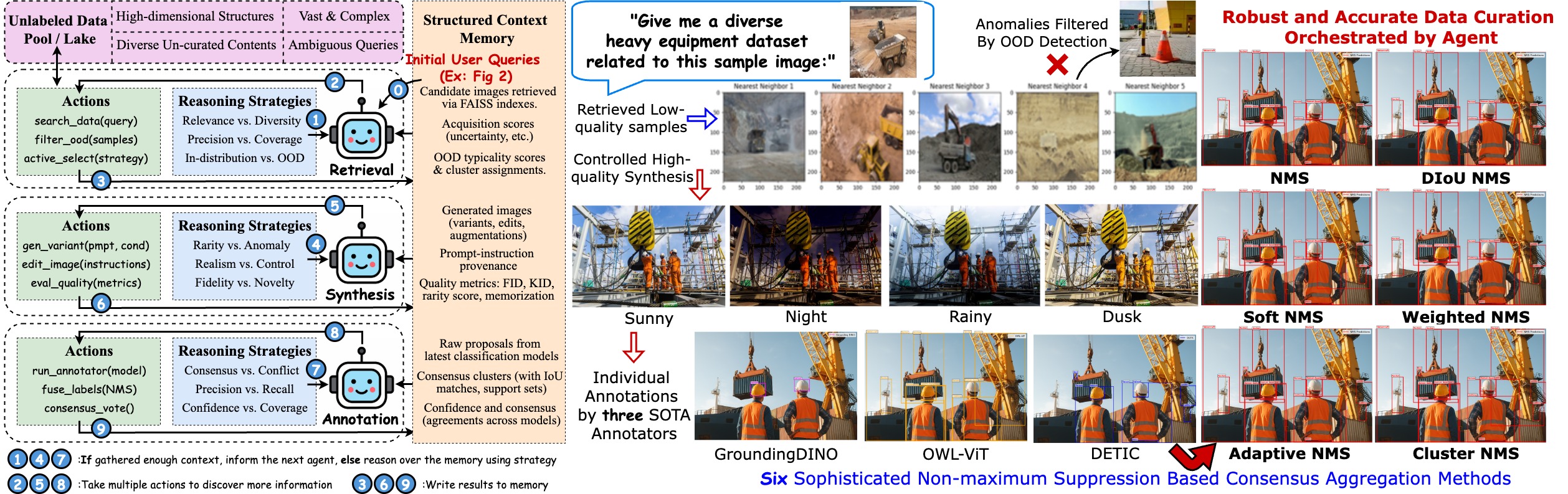}
    \vspace{-3ex}
    \caption{Overview of \projname{}. \textbf{Left:} The system operates in three phases—Retrieval, Synthesis, and Annotation—with a Structured Context Memory storing candidates, synthetic variants, weak labels, and quality signals to guide reasoning. Retrieval balances relevance vs. diversity with active learning and OOD filtering; Synthesis adds rare but realistic data with fidelity and novelty checks; Annotation fuses outputs from multiple detectors via consensus and NMS. \textbf{Right:} an example trajectory shows how retrieved samples, controlled edits, and model proposals are combined into validated annotations, producing a curated dataset that feeds back into the loop.}
    \label{fig:placeholder}
\end{figure*}

This data curation challenge presents several intertwined difficulties that require sophisticated reasoning and tool coordination. Firstly, data sources are often massive, heterogeneous data lakes like \textit{LAION}~\cite{schuhmann2022laion} or \textit{DataComp}~\cite{gadre2023datacomp}, containing millions of irrelevant or low-quality images that must be intelligently filtered~\cite{birhane2021multimodal}. Secondly, real-world applications demand robustness to rare events (e.g., adverse weather, unusual object poses)~\cite{hendrycks2021many, koh2021wilds}, which are inherently absent or underrepresented in training sets—a problem especially acute in data-scarce domains like medical imaging~\cite{larrazabal2020gender} or industrial inspection~\cite{bergmann2019mvtec}. Finally, the annotation process itself is ambiguous; manual labeling is slow and expensive~\cite{su2012crowdsourcing}, while automated methods using different foundation models often produce conflicting labels~\cite{papadopoulos2017training}, introducing significant noise if naively combined~\cite{northcutt2021confident}.

Overcoming these complexities, a daunting task even for human teams, requires a new type of agent -- one that can research the data landscape, synthesize information, and reason about trade-offs. Inspired by these challenges, we propose \textbf{\ul{the first deep research agent for vision data curation, called \textsc{Labeling Copilot}}}. As shown in Figure \ref{fig:placeholder}, our agent operates in a continuous loop: (1) Discovery: Starting with a high-level query, the agent uses a calibrated retrieval tool to find relevant data using techniques from active learning~\cite{settles2009active} and out-of-distribution detection~\cite{yang2021generalized}. (2) Synthesis: If the dataset lacks diversity, it uses a controllable synthesis tool leveraging instruction-following diffusion models~\cite{nichol2021glide, saharia2022photorealistic} to generate images covering specific edge cases. (3) Annotation \& Filtering: The agent orchestrates a suite of foundation models to generate candidate labels, builds a consensus for a single high-quality annotation using techniques inspired by ensemble learning~\cite{dietterich2000ensemble}, and filters out low-confidence samples. This curated data is then used to fine-tune a target model, allowing the agent to repeat the cycle and address model weaknesses iteratively. Overall, our primary contributions are as follows:

\vspace{1ex}
\noindent\circled{1} \textbf{An Agentic Framework for Data Curation:} We design and implement a novel agentic system that unifies data retrieval, synthesis, and annotation into a cohesive, goal-driven workflow. Unlike rigid pipelines, our agent intelligently orchestrates these tools, utilizing quality signals passed between modules to determine whether to dynamically find more real data, synthesize novel examples for rare cases, or fuse annotations to enhance label quality.

\vspace{1ex}
\noindent\circled{2} \textbf{A Scalable, Calibrated Retrieval Tool:} We design an agent-controlled retrieval module that combines active learning strategies~\cite{sener2017active, ash2019deep} with Out-of-Distribution (OOD) detection~\cite{ganguly2025forte}. By reformulating classical active learning algorithms within an FAISS-based approximate nearest neighbor framework, the tool delivers reliable quality signals at scale, handling datasets with over $10$ million samples. This enables the agent to efficiently surface the most informative data for downstream model training.

\vspace{1ex}
\noindent\circled{3} \textbf{A Language-Controllable Synthesis Tool:} We empower the agent with a unified synthesis tool that generates targeted, realistic data from natural language instructions. By combining instruction-following diffusion models~\cite{brooks2023instructpix2pix, zhang2023adding} with multimodal large language models, the agent can create fine-grained semantic variations (e.g., ``a car at night in the rain'') to systematically plan and address data gaps while preserving object identity.

\vspace{1ex}
\noindent\circled{4} \textbf{A Multi-Tool Consensus Annotation Strategy.} We propose a consensus-based annotation strategy where the agent orchestrates multiple foundation models (e.g., DETIC~\cite{zhou2022detecting}, GroundingDINO~\cite{liu2023grounding}) as independent ``expert" tools. The agent then employs sophisticated aggregation mechanisms, such as Soft-NMS~\cite{bodla2017soft} and Weighted-NMS~\cite{zhou2019iou}, to synthesize their diverse outputs, transforming noisy weak labels into high-quality pseudo-labels that are more robust than any single model could produce.

\begin{figure}
    \centering
    \includegraphics[width=1\linewidth]{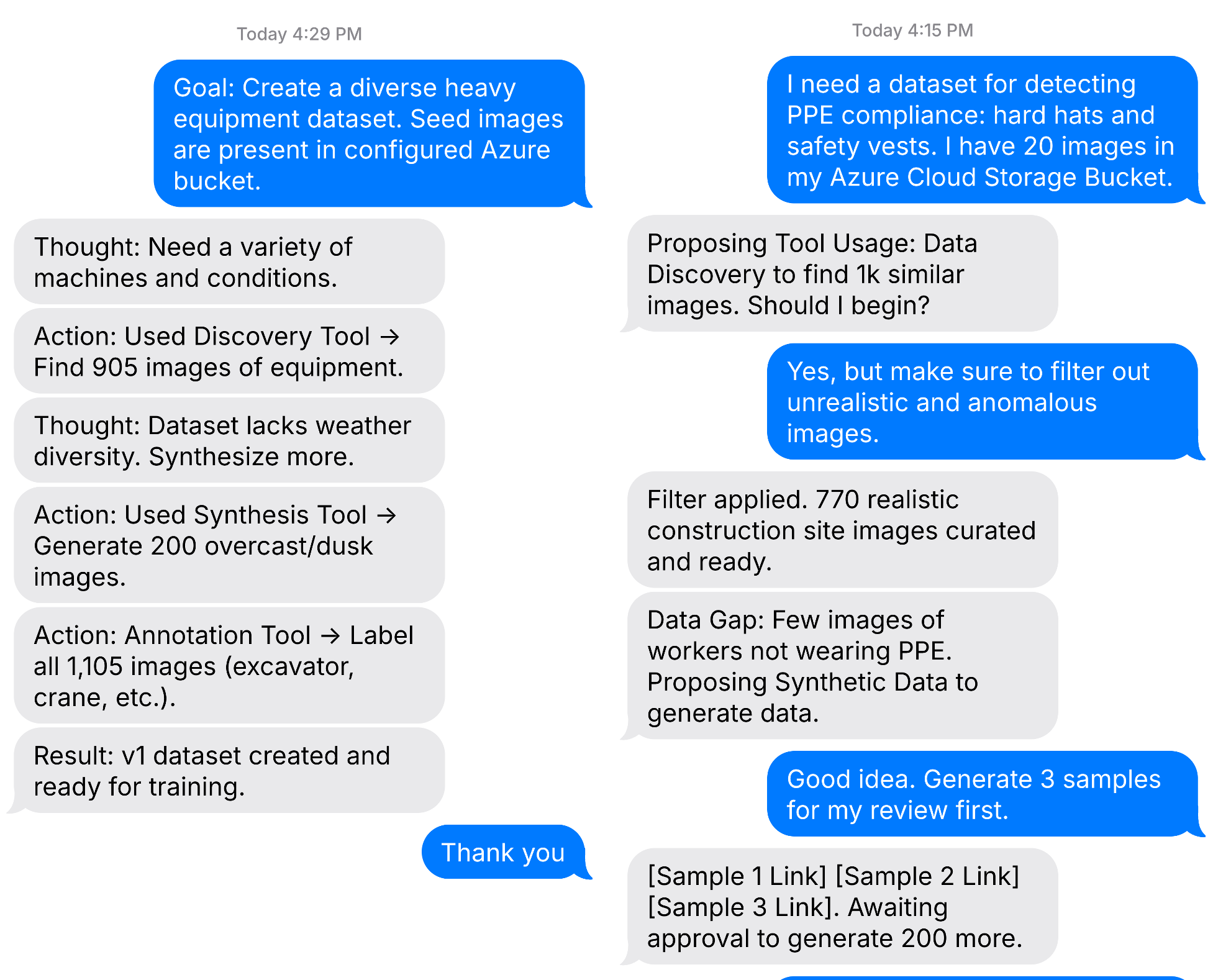}
    \caption{Illustrative demonstration two trajectories for dataset curation: a fully autonomous agent performing discovery, synthesis, and annotation (left), and a human-in-the-loop session where a user guides the agent's actions (right).}
    \label{fig:placeholder}
\end{figure}

\section{Related Work}

\projname{} addresses the critical challenge of automated dataset curation by integrating advances in data acquisition, data labeling, and data quality improvement. While prior efforts tackle these aspects in isolation, our agentic framework uniquely orchestrates them into a unified system that can reason about trade-offs and adapt strategies based on data characteristics and quality signals.

\vspace{1ex}
\noindent\textbf{Data Acquisition and Discovery:}
Effective machine learning relies on acquiring sufficient, relevant data. Traditional approaches span \emph{discovery}, \emph{augmentation}, and \emph{generation}~\cite{roh2019survey}.  
Discovery efforts focus on indexing and sharing large repositories: early platforms such as Google Fusion Tables and \textit{Kaggle} democratized dataset access, while enterprise systems like \textit{Datahub} provide versioning and metadata management. \textit{GOODS} scales further by cataloging billions of datasets for efficient search and provenance. Despite these advances, identifying task-relevant visual data remains difficult due to the complexity of semantic similarity.  
Data augmentation has evolved from simple transformations to advanced mixing strategies, including \textit{Mixup}~\cite{zhang2017mixup}, \textit{CutMix}~\cite{yun2019cutmix}. Automated policy search methods like \textit{AutoAugment}~\cite{cubuk2019autoaugment} and \textit{RandAugment}~\cite{cubuk2020randaugment}. Recent transformer-based approaches such as \textit{TransMix}~\cite{chen2022transmix} and \textit{MixPro}~\cite{zhao2023mixpro} exploit attention maps for more intelligent mixing. Yet, these methods cannot produce semantic, photorealistic modifications (e.g., altering weather or lighting).  
Generative approaches address this gap. While platforms like Amazon Mechanical \textit{Turk} support manual data creation, modern GANs~\cite{goodfellow2014generative} and diffusion models~\cite{ho2020denoising,rombach2022ldm} enable high-quality synthetic data for images, text, and tabular domains~\cite{trabucco2023effective}. However, diffusion-based data often underperforms real samples~\cite{burg2023image}, underscoring the need to integrate retrieval and generation intelligently.

\vspace{1ex}
\noindent\textbf{Data Labeling and Annotation:}
Acquisition must be coupled with effective annotation strategies. Approaches range from fully supervised to weakly supervised, each with accuracy–scalability trade-offs.  
Semi-supervised learning leverages small labeled sets to expand larger unlabeled corpora, with classical methods including self-training~\cite{yarowsky1995unsupervised}, co-training~\cite{blum1998combining}, and tri-training~\cite{zhou2005tri}, which underpin modern pseudo-labeling.  
Active learning prioritizes informative samples via uncertainty sampling~\cite{lewis1994heterogeneous}, query-by-committee~\cite{seung1992query}, and geometric strategies such as K-Center Greedy~\cite{sener2017active}. Recent work explores adaptive multi-armed bandit frameworks~\cite{bouniot2023towards} to combine strategies dynamically.  
Weak supervision exploits noisy sources at scale. Data programming (e.g., \textit{Snorkel}~\cite{ratner2017snorkel}) aggregates labeling functions probabilistically, while multi-instance learning (MIL)~\cite{dietterich1997solving} infers instance-level labels from coarser supervision. Although powerful, these methods introduce noise that requires careful denoising.

\vspace{1ex}
\noindent\textbf{Foundation Models for Vision Tasks:}
Large vision-language foundation models now provide strong automated labeling capabilities. \textit{CLIP}~\cite{radford2021learning} supports zero-shot classification, \textit{DETIC}~\cite{zhou2022detecting} and \textit{GroundingDINO}~\cite{liu2023grounding} enable open-vocabulary detection, and \textit{SAM}~\cite{kirillov2023segment} delivers universal segmentation. However, their outputs often conflict, motivating consensus-based fusion. Recent multitask models such as \textit{Florence-2}~\cite{xiao2024florence}, trained on FLD-5B, highlight the promise of unified architectures, yet curating such datasets remains a major bottleneck requiring intelligent filtering.  
Beyond acquisition, data quality improvement is crucial when new data is limited. Systems such as \textit{HoloClean}~\cite{rekatsinas2017holoclean}, \textit{ActiveClean}~\cite{krishnan2016activeclean}, and \textit{BoostClean}~\cite{schelter2018automating} demonstrate structured-data cleaning at scale. In computer vision, filtering typically relies on non-maximum suppression and its variants to reduce redundant detections, though these are usually applied per model rather than orchestrated across multiple detectors.

\vspace{1ex}
\noindent\textbf{Positioning of Our Work:}
Existing approaches address individual steps in data curation but lack intelligent orchestration. \projname{} uniquely integrates retrieval, synthesis, and consensus annotation into an agentic framework that reasons about data quality, adapts strategies iteratively, and makes principled trade-offs. To our knowledge, this is the first work to frame dataset curation as an agentic research process, transforming disconnected tools into a cohesive, goal-driven system.

\section{Methodology: An Agent-Driven Toolkit}

The \projname{} framework is built around a central orchestrator agent that intelligently manages a suite of specialized tools, qualifying as a deep research agent under the formal definition established by recent literature \cite{java2025characterizing} through its high search intensity (processing millions of information units via scalable active learning algorithms), reasoning intensity across all three complexity dimensions (finding information units via sophisticated retrieval strategies, processing them through multi-model consensus mechanisms, and combining conflicting outputs via advanced NMS variants), and dynamic agent orchestration where the central coordinator intelligently selects and sequences specialized tools based on data characteristics and quality signals, with the capability to iteratively adapt strategies (e.g., invoking synthesis when discovery yields insufficient diversity) throughout the curation workflow. This section details the core tools the agent uses for data discovery, synthesis, and annotation.

\subsection{The Data Discovery Tool}

The agent's first task is to source relevant data from a large, unlabeled data pool $\mathcal{U}$. It employs the data discovery , which is designed to select an informative batch $\mathcal{B}$ for labeling. To enable efficient similarity search, all data points in $\mathcal{U}$ are pre-indexed using FAISS \cite{ douze2024faiss}. The workflow consists of two main stages: selecting a batch of candidate samples using a scalable active learning (AL) strategy, and then filtering out-of-distribution (OOD) samples from that batch.

\vspace{1ex}
\noindent\textbf{Stage 1: Scalable Active Learning for Candidate Selection} Canonical active learning (AL) algorithms are fundamentally incompatible with the scale of modern datasets. Their reliance on operations that scan the entire unlabeled data pool—such as full distance matrix computations or exhaustive model inference—introduces a computational bottleneck that is untenable in a big data context. Our methodology systematically dismantles this bottleneck by reformulating core AL strategies around the sub-linear time complexities of Approximate Nearest Neighbor (ANN) search. We utilize the FAISS library not only for search but also as a computational engine to replace brute-force operations with efficient, index-based approximations.

We anchor our methodology in FAISS, a library purpose-built for vector similarity search on massive-scale datasets. Rather than using it as a simple search tool, we employ its indexing structures as a computational substrate to overcome specific scalability challenges in active learning. Our strategic choices include:

\begin{itemize}
    \item \textit{Tackling Search Complexity with Inverted File Systems (IVF):} To avoid linear scans, we partition the vector space using \texttt{IndexIVF} structures. The dataset is divided into \texttt{nlist} Voronoi cells, and at query time, the search is constrained to a small subset, \texttt{nprobe}, of these cells. This reduces search complexity from $\mathcal{O}(N)$ to approximately $\mathcal{O}(\frac{\texttt{nprobe}}{\texttt{nlist}} \cdot N)$, forming the basis of our localized sampling strategies.

    \item \textit{Managing Memory Footprint with Product Quantization (PQ):} For datasets exceeding main memory capacity, we employ \texttt{IndexIVFPQ} to compress vectors. By decomposing vectors into sub-vectors and quantizing each independently, PQ dramatically reduces the per-vector memory cost from $4d$ bytes to as few as 8 or 16 bytes, making billion-scale in-memory active learning feasible.

    \item \textit{Ensuring High-Recall Search with HNSW Graphs:} In regimes where high search accuracy is paramount, we utilize Hierarchical Navigable Small World (\texttt{IndexHNSW}) graph indexes. HNSW provides superior speed-recall trade-offs compared to IVF-based methods and can serve either as a primary index or as a powerful coarse quantizer for a hybrid IVF system (e.g., \texttt{IVF65536\_HNSW32,Flat}).
\end{itemize}

The efficacy of our entire framework hinges on the synergistic use of two fundamental FAISS operations: \texttt{search()} for efficient candidate localization and \texttt{reconstruct\_batch()} for retrieving the full-precision vectors required for model training. This combination transforms the FAISS index from a passive search structure into an active engine for scalable data selection.

\vspace{1ex}
\noindent\textbf{Scalable Reformulation of Active Learning Algorithms:} With this tool, our primary contribution is the redesign of classic AL algorithms to operate on small, intelligently-selected data subsets rather than the entire unlabeled pool.

\textit{K-Center Greedy via Candidate Subsampling:}
The K-Center Greedy algorithm seeks to select points that maximally cover the feature space. Its brute-force implementation, however, requires iteratively computing distances from every unlabeled point to the growing set of labeled centers, a process with complexity that scales prohibitively with the dataset size $|\mathcal{U}|$.

To overcome this, we implement an approximate version that operates on a fixed-size candidate pool $\mathcal{U}_c \subset \mathcal{U}$ of size $N_c$. This pool is randomly sampled from the unlabeled set at the beginning of each selection round. The greedy selection logic then proceeds as normal, but is confined to this computationally manageable subset, by using \texttt{faiss.pairwise\_distances}, we leverage optimized BLAS routines for the core distance calculations. This reformulation bounds the complexity of selecting a batch of size $B$ to $\mathcal{O}(B \cdot N_c \cdot |\mathcal{L}| \cdot d)$, making the algorithm's runtime independent of the total dataset size $|\mathcal{U}|$ and dependent only on the configurable pool size $N_c$.

\textit{Localized Acquisition for Uncertainty and Representative Sampling:}
A broad class of powerful AL strategies, including margin sampling, entropy sampling, and representative sampling, is predicated on first identifying points of high uncertainty. The common bottleneck across all these methods is the need to perform model inference over every point in $\mathcal{U}$ to compute these uncertainty scores.

Our solution is to replace this exhaustive inference with an efficient, ANN-driven localization. We hypothesize that the most informative samples for a model are located in the feature space neighborhood of the data it has already been trained on. We formalize this as a generalized localized acquisition strategy, detailed in Algorithm~\ref{alg:localized_sampling}.

This strategy creates a compact ``micro-pool" $\mathcal{N}_u$ of promising candidates, drastically reducing the scope of expensive downstream computations. For \textit{Margin Sampling}, we compute decision boundary distances for points within $\mathcal{N}_u$. For more complex \textit{Representative Sampling} strategies, we first identify the most uncertain points within $\mathcal{N}_u$ and then apply a fast clustering algorithm (e.g., \texttt{MiniBatchKMeans}) to this much smaller set to select a diverse batch of medoids. In all cases, the cost of inference is reduced from $\mathcal{O}(|\mathcal{U}|)$ to $\mathcal{O}(K_s)$, where the neighborhood size $K_s$ is a tunable hyperparameter orders of magnitude smaller than $|\mathcal{U}|$.

\begin{figure}
    \centering
    \includegraphics[width=1\linewidth]{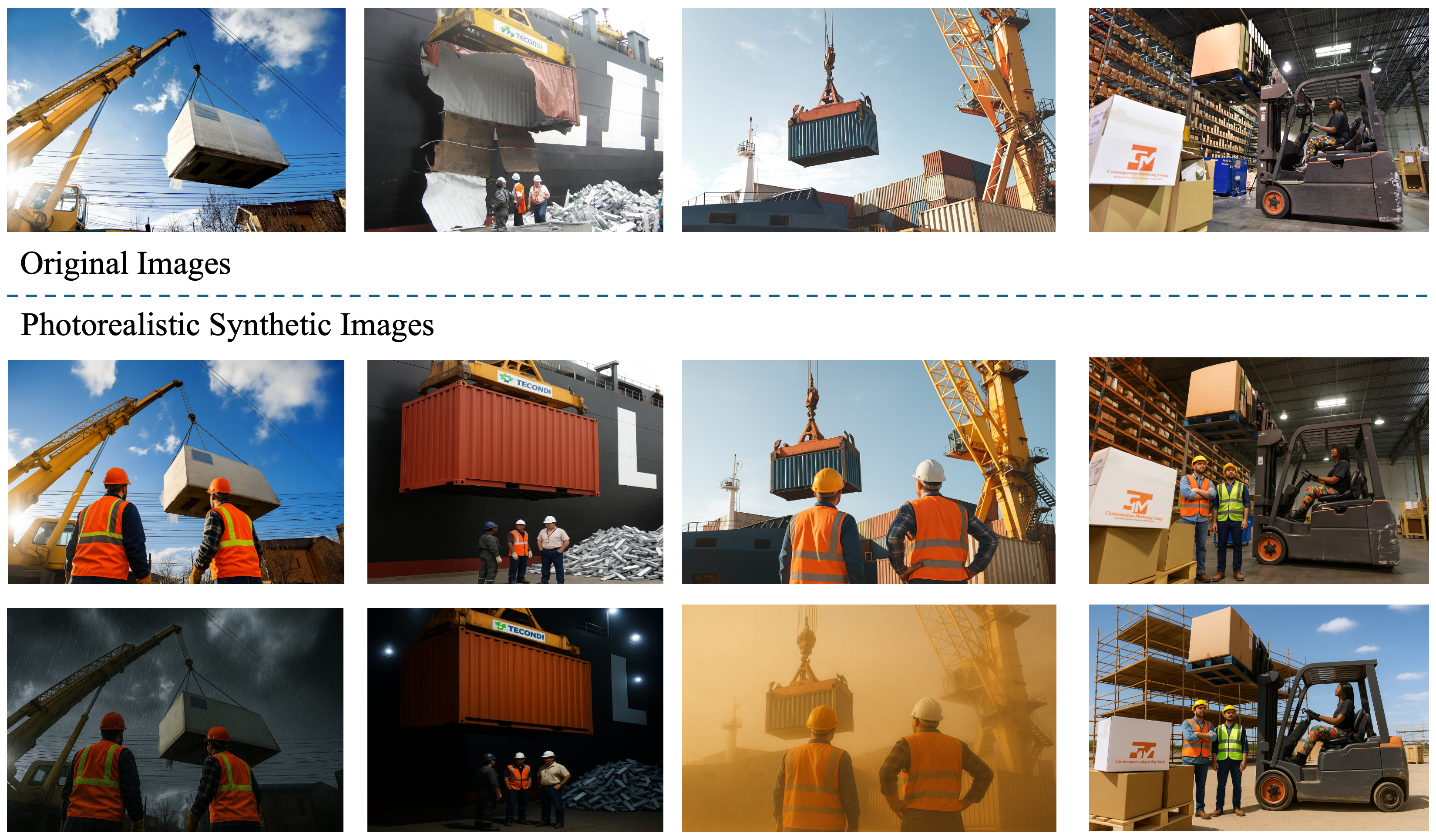}
    \vspace{-3ex}
    \caption{The \projname{}'s synthesis  generates diverse, photorealistic variants (middle and bottom rows) from original industrial images (top row). This capability is critical for creating training data for workplace hazards that are unethical or impossible to capture in reality, such as workers standing under a suspended load (a serious OSHA violation). The agent can also alter environmental conditions (e.g., adding rain or dust) and context (e.g., moving an indoor scene outdoors) to create data for rare scenarios, thereby systematically improving the robustness of safety detection computer vision models.}
    \label{fig:placeholder}
\end{figure}

\begin{algorithm}[t]
\caption{Generalized Localized Active Learning Framework}
\label{alg:localized_sampling}
\begin{algorithmic}[1]
\Require FAISS index $\mathcal{I}$, Labeled set IDs $\text{ids}(\mathcal{L})$, Batch size $B$, Neighborhood size $K_s$.
\State $\mathbf{V}_\mathcal{L} \leftarrow \mathcal{I}.\text{reconstruct\_batch}(\text{ids}(\mathcal{L}))$. \Comment{Retrieve labeled vectors}
\State $\mathbf{q} \leftarrow \text{ComputeCentroid}(\mathbf{V}_\mathcal{L})$. \Comment{Define search query (e.g., centroid)}
\State $\text{ids}(\mathcal{N}) \leftarrow \mathcal{I}.\text{search}(\mathbf{q}, K_s)$. \Comment{Find local neighborhood via ANN search}
\State $\text{ids}(\mathcal{N}_u) \leftarrow \text{ids}(\mathcal{N}) \setminus \text{ids}(\mathcal{L})$. \Comment{Isolate unlabeled candidates}
\State $\mathbf{V}_{\mathcal{N}_u} \leftarrow \mathcal{I}.\text{reconstruct\_batch}(\text{ids}(\mathcal{N}_u))$. \Comment{Retrieve candidate vectors}
\State Train model $\mathcal{M}$ on $(\mathbf{V}_\mathcal{L}, \mathbf{y}_\mathcal{L})$.
\State Compute acquisition scores for all $\mathbf{v}_i \in \mathbf{V}_{\mathcal{N}_u}$ using $\mathcal{M}$.
\State Select batch $\mathcal{B}$ of size $B$ from $\text{ids}(\mathcal{N}_u)$ based on scores.
\State \Return $\mathcal{B}$.
\end{algorithmic}
\end{algorithm}

\vspace{1ex}
\noindent\textbf{Stage 2: Filtering Spurious Samples with OOD Detection:} 
To enforce distributional consistency and ensure that the active learning process focuses on semantically relevant samples, we introduce a probabilistic filtering stage after candidate retrieval. This step is designed to reject statistical outliers that may be returned by the ANN search, particularly from sparse regions of the embedding space. We implement this filter by adapting the Forte typicality estimation framework~\cite{ganguly2025forte}, which operates in a learned self-supervised representation space. The process consists of two primary stages: manifold modeling and probabilistic filtering.

First, we model the high-dimensional manifold of the known in-distribution data, represented by the feature set $\mathbf{V}_\mathcal{L}$ from the labeled pool $\mathcal{L}$. We employ a Gaussian Mixture Model (GMM) for this task, chosen for its ability to capture complex, multi-modal data structures inherent in real-world visual classes. The GMM parameters, $\Theta = \{\pi_k, \boldsymbol{\mu}_k, \boldsymbol{\Sigma}_k\}_{k=1}^K$, representing the mixture weights, means, and covariances, are estimated from $\mathbf{V}_\mathcal{L}$ via the Expectation-Maximization (EM) algorithm. This yields a probabilistic model of the in-distribution data density:
$p(\mathbf{x} | \Theta) = \sum_{k=1}^{K} \pi_k \mathcal{N}(\mathbf{x} | \boldsymbol{\mu}_k, \boldsymbol{\Sigma}_k)$

Second, for each retrieved candidate vector $\mathbf{x}'$, we use the fitted GMM to perform probabilistic filtering. Our innovation here is to derive a continuous typicality score rather than a discrete classification. We calculate the posterior probability, or responsibility, of each mixture component $k$ for generating the sample $\mathbf{x}'$:
$\gamma_k(\mathbf{x}') = P(k | \mathbf{x}', \Theta) = \frac{\pi_k \mathcal{N}(\mathbf{x}' | \boldsymbol{\mu}_k, \boldsymbol{\Sigma}_k)}{p(\mathbf{x}' | \Theta)}$
The final typicality score, $S(\mathbf{x}')$, is the maximum responsibility across all components, $S(\mathbf{x}') = \max_{k} \gamma_k(\mathbf{x}')$. This score measures how well $\mathbf{x}'$ aligns with the densest regions of the learned in-distribution manifold. A sample is accepted into the final batch for labeling only if its score surpasses a predefined threshold, $S(\mathbf{x}') \ge \tau$. This mechanism acts as a robust semantic guardrail, preventing the labeling budget from being expended on noisy or irrelevant outliers and thereby promoting a more efficient and stable active learning loop.

\subsection{The Synthetic Data Tool}

When the agent determines that the existing data is insufficient (e.g., lacks diversity or rare cases), it employs the synthetic data  to generate new, high-quality images.


\vspace{1ex}
\noindent\textbf{Step 1: Choosing a Generation Technique:} The agent selects the best synthesis method based on the complexity of the desired image. 

\begin{itemize}
    \item For simple scenes, the agent uses state-of-the-art captioning models (GPT-4, BLIP, Florence 2) to generate rich text prompts, which are then fed to Stable Diffusion, DALL-E.
    \item For modifying natural scenes, the agent can use the original image as a strong conditioning reference for an image-to-image Stable Diffusion model, setting a low strength value to generate realistic variations.
    \item For complex or niche domains, the agent uses GPT-4o to generate a large set of specific editing instructions for a single image (e.g., ``add rain," ``make it nighttime"). It then executes these instructions using a prompt-based image editing model, such as InstructPix2Pix. We achieve the highest quality synthetic images using natively multimodal large language models like GPT-Image-1.
\end{itemize}

\vspace{1ex}
\noindent\textbf{Step 2: Evaluating Synthetic Data Quality:} Before adding synthetic data to the dataset, the agent rigorously evaluates its quality using a comprehensive suite of metrics to assess four key criteria: fidelity, diversity, rarity, and memorization.\cite{stein2023exposing}

\begin{itemize}
    \item \textit{Fidelity \& Diversity:} The agent computes metrics like Fréchet Inception Distance (FID), Kernel Inception Distance (KID), precision, recall, density, and coverage in the DinoV2 representation space.
    \item \textit{Memorization:} When using fine-tuned models, the agent checks for overfitting and memorization using metrics such as Percentage of authentic samples (AuthPCT) and Feature Likelihood Score (FLS).
    \item \textit{Automated Selection:} The agent treats these normalized scores as logits, allowing it to use top-k or top-p sampling to automatically select the best-performing models and prompts for generation, with human experts validating the final hyperparameters.
\end{itemize}

\subsection{The Annotation Tool}

Once a batch of data is curated, the agent employs the annotation tool to generate precise and reliable labels. We present a high-throughput weak supervision framework designed for HPC environments, which transforms vast, unlabeled image corpora into high-quality annotated datasets. The methodology is architected as a two-phase, massively parallel pipeline: (1) weak label generation from a heterogeneous ensemble of foundation models, and (2) consensus-based fusion to distill a robust ground truth from these noisy, multi-source predictions.

\vspace{1ex}
\noindent\textbf{Phase 1: Weak Supervision via a Heterogeneous Ensemble:}
The foundational principle of our framework is to leverage the complementary strengths of diverse, open-vocabulary object detection models to generate a rich set of weak labels. The heterogeneity of the model architectures is a critical design choice, as it promotes uncorrelated error modes, which are essential for effective downstream aggregation and error correction. Our programmatic annotator ensemble comprises models representing distinct architectural paradigms:

\begin{itemize}
    \item \textit{DETIC:} A CLIP-driven detector that excels at zero-shot generalization by dynamically embedding a user-defined text vocabulary into its classification head.
    \item \textit{GroundingDINO:} A sophisticated encoder-decoder transformer that performs language-to-vision grounding, enabling it to detect objects specified by complex, free-form text prompts.
    \item \textit{OWL-ViT:} A Vision Transformer (ViT) architecture adapted for open-vocabulary detection, providing a distinct feature extraction and localization mechanism.
\end{itemize}

For a given image $I \in \mathcal{D}$ from a dataset corpus and a shared target vocabulary $\mathcal{V}$, each model $M_i$ in the ensemble $\mathcal{M} = \{M_1, \dots, M_N\}$ generates an independent set of proposals $P_i$. Each proposal $p_{ij} \in P_i$ is a tuple $(b_{ij}, c_{ij}, s_{ij})$, consisting of a bounding box $b_{ij} \in \mathbb{R}^4$, a class label $c_{ij} \in \mathcal{V}$, and a model-specific confidence score $s_{ij} \in [0, 1]$.

\textit{HPC Implementation Strategy:}
The annotation of each image by each model is a stateless, independent task. This allows for near-linear scalability on a distributed computing cluster, where each image-model pair can be scheduled as a separate job. To facilitate this and ensure modularity, the output for each image is a collection of standardized PASCAL VOC XML files—one for each model. This use of a file-based intermediate representation decouples the generation and fusion stages, a critical feature for managing complex, large-scale workflows and enhancing fault tolerance.

\vspace{1ex}
\noindent\textbf{Phase 2: Consensus-Based Annotation Fusion:}
The weak labels produced in Phase 1 are numerous, often conflicting, and contain significant noise. The fusion phase distills these raw proposals into a single, high-confidence annotation set for each image through a principled, voting-based algorithm.

\textit{Consensus Set Generation via Support-Based Matching:}
Unlike traditional NMS, which operates on proposals from a single model, our fusion logic must reconcile predictions from multiple, disparate sources. For each class $c \in \mathcal{V}$, we aggregate all proposals $\{p_j\}$ from all models. For each proposal $p_j$, we compute its \textit{support set} by identifying the best-matching proposal (if any) from each of the other models $M_i$ based on an Intersection over Union (IoU) threshold, $\tau_{iou}$. The collection of a proposal and its supporting matches forms a consensus cluster.

From each cluster, we derive a fused bounding box $b^*$ by averaging the coordinates of all constituent proposals. A consensus confidence score, $\mathcal{S}$, is then computed. This score is not a simple average of model scores but rather a measure of inter-model agreement, defined as the proportion of unique models in the ensemble that support the cluster:

$$ \mathcal{S}(C_k) = \frac{|\{ M_i \in \mathcal{M} \mid \exists p \in P_i \text{ s.t. } p \in C_k \}|}{N} $$

where $C_k$ is the set of proposals in the $k$-th cluster. This mechanism intrinsically up-weights detections that are consistently identified across different model architectures. The output of this step is a refined set of candidate annotations, $A^* = \{(b_k^*, c_k, \mathcal{S}_k)\}_{k=1}^K$.

\textit{Finalization via Configurable Non-Maximal Suppression:}
The candidate set $A^*$ represents high-agreement detections but may still contain spatial overlaps. We employ a final, configurable filtering module based on advanced NMS variants to resolve these conflicts. The consensus score $\mathcal{S}$ is used as the primary sorting criterion. Our framework integrates several strategies to handle diverse object distributions:
\begin{itemize}
    \item \textit{DIoU-NMS:} We utilize a metric that penalizes for the distance between box centers in addition to IoU. This is our default, as it yields more robust suppression for occluded or proximate objects by considering the centrality of the detections.
    \item \textit{Soft-NMS:} For dense scenes, we employ a non-destructive suppression where the scores of overlapping boxes are decayed as a Gaussian function of their IoU, preserving plausible but overlapping detections.
\end{itemize}
The final output of this two-phase pipeline, $A_{final} = \text{NMS}(A^*, \tau_{nms})$, is a single, programmatically-generated annotation file for each image, synthesized from a robust process of evidence aggregation, voting, and refinement.

\section{Results and Evaluation}

Evaluating a long-running, autonomous agent like \projname{} presents a non-trivial benchmarking challenge. Its purpose is to generate datasets, a task with a vast and open-ended action space. Consequently, a traditional end-to-end evaluation based on a single, fixed downstream task would fail to capture the agent's general-purpose utility across diverse domains. To address this, we adopt \textbf{a data-centric evaluation protocol:} Instead of measuring a single downstream outcome, we perform a rigorous, component-wise analysis focused on the quality and characteristics of the data artifacts produced at each stage of the agentic workflow. This approach allows us to thoroughly validate the performance of \projname{}'s core tools on both academic and industry-specific datasets. The following sections present the key findings from our experiments, demonstrating the effectiveness of each component.

\subsection{Protocol for Scalability and Efficacy Evaluation}

\begin{figure}[htbp]
    \centering
    \includegraphics[width=1\linewidth]{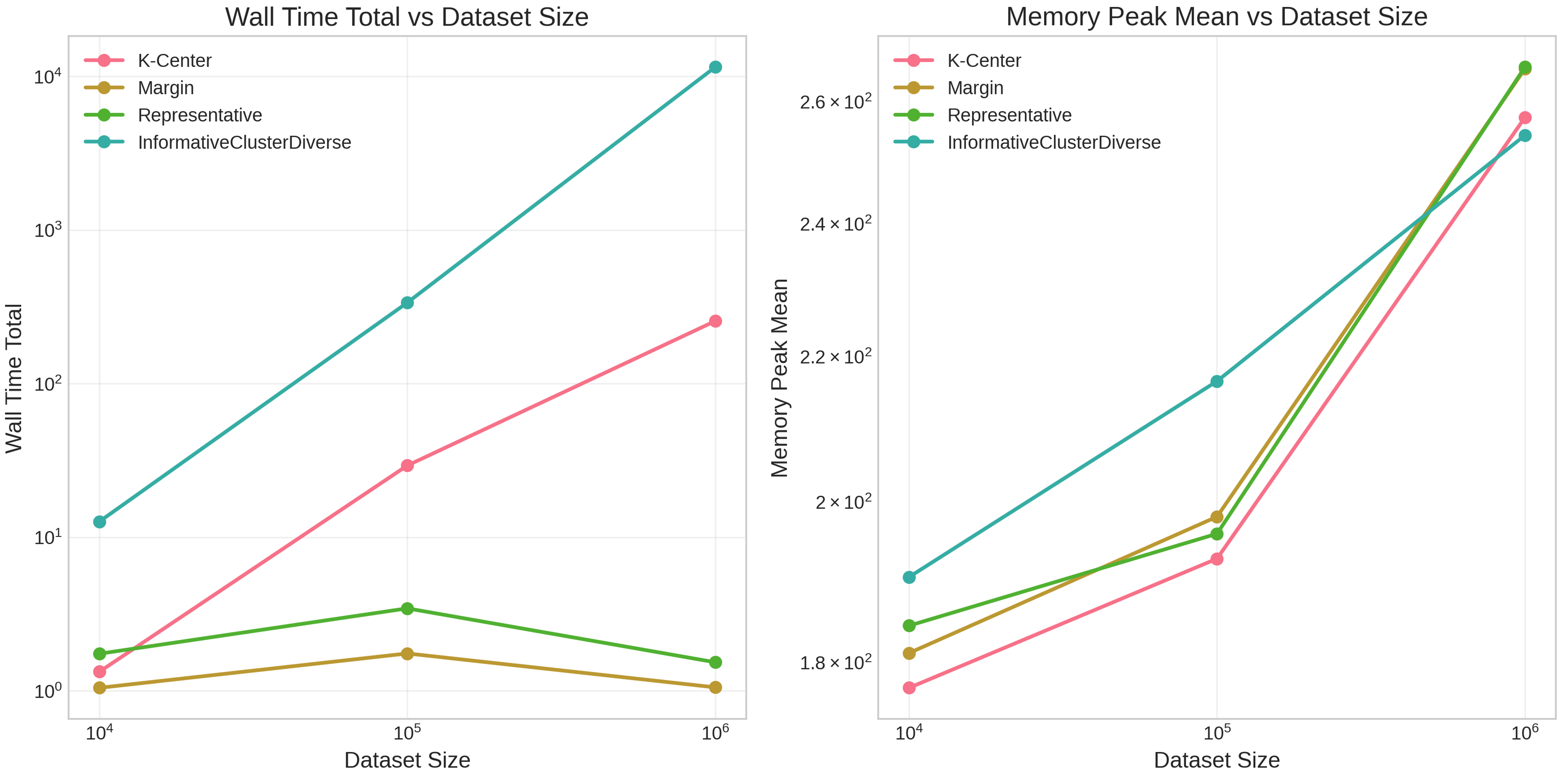}
    \vspace{-2ex}
    \caption{Computational scaling behavior of active learning algorithms across dataset sizes from 10K to 1M samples. The left panel shows wall-clock time scaling on log-log axes, revealing fundamental algorithmic complexity differences: InformativeClusterDiverse exhibits super-linear growth (15s to 10,000s), K-Center demonstrates favorable sub-linear scaling (1s to 250s), while Margin and Representative show flat profiles. The right panel displays memory consumption scaling, with all methods following similar power-law growth patterns that converge at large scales, suggesting memory bottlenecks shift from vector storage to auxiliary computational structures. The parallel scaling curves indicate that algorithmic choice, rather than data structure optimization, becomes the primary determinant of computational feasibility at production scales. Note that these measurements aggregate across different index architectures (Flat, IVF, HNSW) that were varied with dataset size.}
    \label{fig:placeholder}
\end{figure}

We designed a rigorous experimental protocol to evaluate the performance of our scalable AL framework under diverse conditions. 
The independent variables include: (i) \textit{dataset scale}, ranging from $N=10^4$ to $10^8+$ samples with dimensionality $d \in \{64,128,256\}$; 
(ii) \textit{index architecture}, which progresses with scale from \texttt{Flat} to \texttt{IVF,Flat} and ultimately to compressed \texttt{IVF,PQ} structures; and 
(iii) \textit{AL hyperparameters}, specifically the candidate pool size $N_c$ for K-Center and the local neighborhood size $K_s$ for localized methods. 
Performance is assessed using two dependent measures: \textit{sample efficiency}, defined as downstream classifier accuracy versus the number of acquired labels, and \textit{computational cost}, measured by wall-clock batch selection time, CPU usage, and peak memory footprint.

Our analysis reveals that algorithmic architecture is the dominant factor in system performance, far outweighing vector database optimization. At the 10M sample scale, the choice of algorithm created a distinct performance hierarchy, while different index architectures all converged to a similar accuracy (0.843-0.851 AUC, $<$1\% variance). The FAISS-optimized K-Center algorithm emerges as the optimal choice, matching the sample efficiency of the more complex InformativeClusterDiverse (both reaching an AUC of 0.90) while being 40 times more computationally efficient (250 vs. 10,000+ seconds). This efficiency is critical, as InformativeClusterDiverse exhibits a debilitating super-linear time complexity, scaling from 15 to over 10,000 seconds between 10,000 and 1 million samples. Simpler methods like Margin and Representative failed to scale, indicated by flat performance curves that suggest systematic experimental failures. Ultimately, these findings establish that for extreme-scale active learning, optimization should prioritize algorithmic robustness, where K-Center provides the best balance of sample efficiency and computational feasibility.

\begin{figure}[htbp]
    \centering
    \includegraphics[width=1\linewidth]{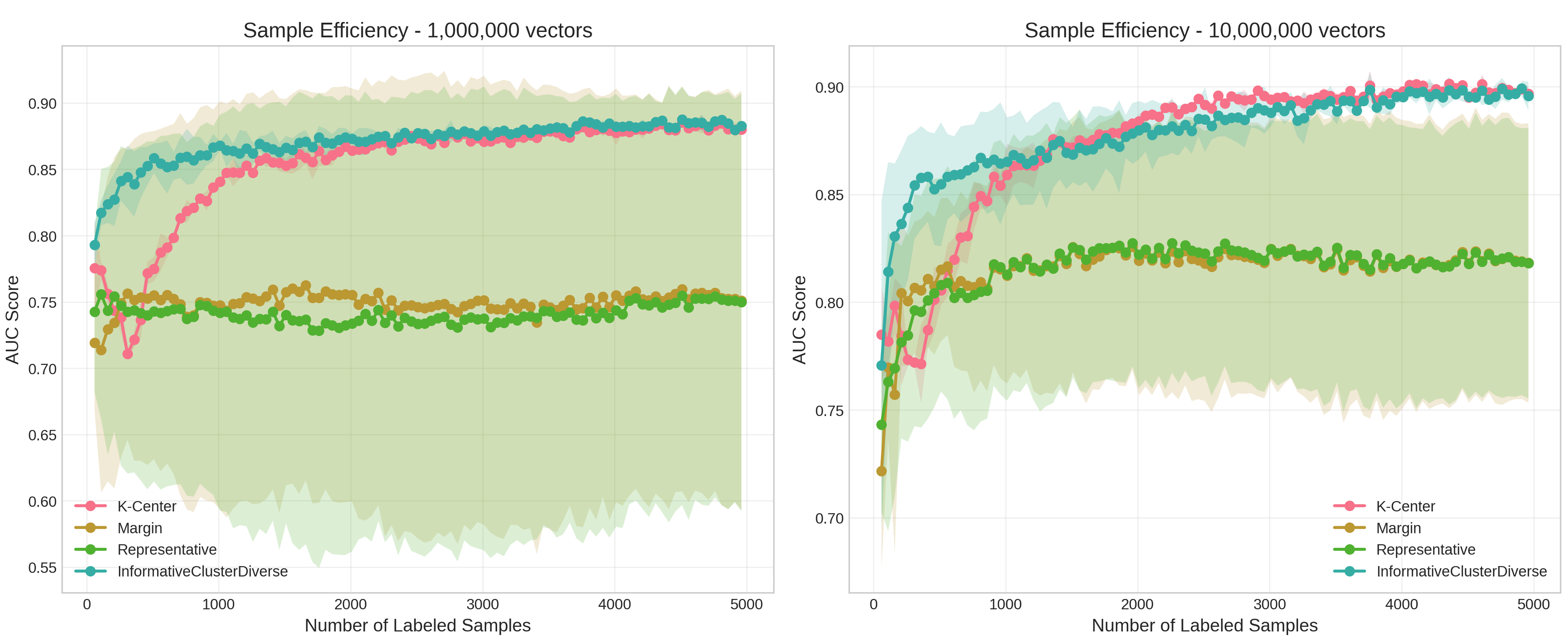}
    \vspace{-3ex}
    \caption{Sample efficiency curves for active learning algorithms at 1M and 10M vector scales. Learning curves display AUC performance as a function of labeled sample count across four sampling strategies: K-Center (diversity-based), Margin (uncertainty-based), Representative (combining uncertainty and diversity), and Informative Cluster Diverse (combining uncertainty, clustering, and diversity). At 1M scale (left), InformativeClusterDiverse and K-Center achieve superior performance (0.87-0.88 AUC), while Margin and Representative plateau at lower performance levels (0.74-0.75 AUC). At 10M scale (right), K-Center matches InformativeClusterDiverse's final performance (both reaching 0.90 AUC) while maintaining faster convergence. The performance gap between sophisticated and lightweight methods widens at extreme scale, with robust algorithms demonstrating consistent improvement throughout the labeling budget while simpler methods exhibit early saturation. Curves represent mean performance across multiple random seeds with standard deviation bands, using synthetic binary classification tasks with 128-dimensional feature vectors.}
    \label{fig:placeholder}
\end{figure}

\subsection{Quantitative Analysis of Annotation Quality}

\begin{table*}[htbp]
\centering
\caption{Comprehensive Annotation Quality Comparison of NMS Methods across COCO, Open Images, and Pascal VOC Datasets}
\label{tab:main_results}
\resizebox{\textwidth}{!}{
\begin{tabular}{lcccccccccccccccccc}
\toprule
\multirow{2}{*}{\textbf{Dataset}} & \multirow{2}{*}{\textbf{Method}} & \textbf{Precision} & \textbf{Recall} & \textbf{F1} & \textbf{mAP} & \textbf{mAP} & \textbf{mAP} & \textbf{Avg} & \textbf{Overlap} & \textbf{Sm\%} & \textbf{Med\%} & \textbf{Lg\%} & \textbf{Imbalance} & \textbf{Avg GT} & \textbf{Avg Pred} & \textbf{Avg Correct} & \textbf{Coverage} & \textbf{Discovered}\\
& & & & & \textbf{@0.5} & \textbf{@0.75} & \textbf{@.5:.95} & \textbf{IoU} & \textbf{Ratio} & \textbf{($<$1\%)} & \textbf{(1-5)\%} & \textbf{($>$5\%)} & \textbf{Ratio} & \textbf{Obj/Img} & \textbf{Obj/Img} & \textbf{Obj/Img} & \textbf{GT \%} & \textbf{New Classes}\\
\midrule
\multirow{6}{*}{\rotatebox{90}{COCO}} 
& NMS & 0.486 & 0.799 & 0.565 & 0.460 & 0.397 & 0.368 & 0.118 & 0.920 & 48.4 & 26.6 & 25.0 & 13325 & 7.43 & 14.20 & 5.24 & 92.4 & 598\\
& Adaptive NMS & \textbf{0.494} & 0.771 & 0.564 & 0.445 & 0.384 & 0.356 & 0.112 & 0.912 & 49.2 & 26.5 & 24.3 & 12753 & 7.43 & 13.43 & 5.06 & 92.4 & 598\\
& Soft NMS & 0.481 & \textbf{0.802} & 0.562 & \textbf{0.461} & \textbf{0.401} & \textbf{0.371} & 0.123 & 0.923 & 48.1 & 26.8 & 25.2 & 13489 & 7.43 & 14.55 & 5.26 & 92.4 & 598\\
& DIoU NMS & 0.485 & 0.799 & 0.565 & 0.460 & 0.397 & 0.368 & 0.119 & 0.921 & 48.3 & 26.7 & 25.0 & 13344 & 7.43 & 14.24 & 5.24 & 92.4 & 598\\
& Weighted NMS & 0.486 & 0.799 & \textbf{0.565} & 0.460 & 0.397 & 0.368 & 0.118 & 0.920 & 48.4 & 26.6 & 25.0 & 13325 & 7.43 & 14.20 & 5.24 & 92.4 & 598\\
& Cluster NMS & 0.488 & 0.762 & 0.554 & 0.449 & 0.392 & 0.362 & \textbf{0.134} & 0.904 & 41.1 & 28.8 & 30.1 & 10178 & 7.43 & 11.93 & 4.62 & 92.4 & 598\\
\midrule
\multirow{6}{*}{\rotatebox{90}{Open Images}} 
& NMS & 0.234 & 0.454 & 0.252 & 0.174 & 0.148 & 0.142 & 0.116 & 0.983 & 45.6 & 26.5 & 27.9 & 65813 & 8.15 & 22.55 & 3.48 & 57.7 & 903\\
& Adaptive NMS & \textbf{0.235} & 0.442 & 0.251 & 0.170 & 0.144 & 0.138 & 0.115 & 0.982 & 46.4 & 26.4 & 27.2 & 64196 & 8.15 & 21.74 & 3.39 & 57.7 & 903\\
& Soft NMS & 0.230 & \textbf{0.456} & 0.249 & \textbf{0.175} & \textbf{0.150} & \textbf{0.143} & \textbf{0.119} & 0.984 & 45.1 & 26.5 & 28.4 & 65877 & 8.15 & 23.27 & 3.51 & 57.7 & 903\\
& DIoU NMS & 0.233 & 0.455 & \textbf{0.252} & 0.174 & 0.148 & 0.142 & 0.117 & 0.983 & 45.6 & 26.5 & 28.0 & 65824 & 8.15 & 22.63 & 3.49 & 57.7 & 903\\
& Weighted NMS & 0.234 & 0.454 & 0.252 & 0.174 & 0.148 & 0.142 & 0.116 & 0.983 & 45.6 & 26.5 & 27.9 & 65813 & 8.15 & 22.55 & 3.48 & 57.7 & 903\\
& Cluster NMS & 0.233 & 0.445 & 0.250 & 0.173 & 0.149 & 0.142 & 0.128 & 0.980 & 39.5 & 27.3 & 33.2 & 46988 & 8.15 & 19.25 & 3.21 & 57.7 & 903\\
\midrule
\multirow{6}{*}{\rotatebox{90}{Pascal VOC}} 
& NMS & 0.404 & \textbf{0.937} & 0.516 & 0.420 & 0.390 & 0.363 & 0.171 & 0.904 & 27.0 & 24.6 & 48.4 & 6562 & 2.29 & 6.13 & 2.03 & 98.0 & 82\\
& Adaptive NMS & \textbf{0.410} & 0.919 & \textbf{0.517} & 0.413 & 0.383 & 0.358 & 0.163 & 0.895 & 27.3 & 24.7 & 48.0 & 6270 & 2.29 & 5.83 & 1.97 & 98.0 & 82\\
& Soft NMS & 0.401 & \textbf{0.939} & 0.513 & \textbf{0.420} & \textbf{0.392} & \textbf{0.365} & 0.176 & 0.907 & 26.8 & 24.7 & 48.4 & 6627 & 2.29 & 6.26 & 2.04 & 98.0 & 82\\
& DIoU NMS & 0.404 & 0.937 & 0.516 & 0.420 & 0.390 & 0.364 & 0.171 & 0.904 & 27.0 & 24.6 & 48.4 & 6566 & 2.29 & 6.15 & 2.03 & 98.0 & 82\\
& Weighted NMS & 0.404 & \textbf{0.937} & 0.516 & 0.420 & 0.390 & 0.363 & 0.171 & 0.904 & 27.0 & 24.6 & 48.4 & 6562 & 2.29 & 6.13 & 2.03 & 98.0 & 82\\
& Cluster NMS & 0.404 & 0.930 & 0.515 & 0.417 & 0.389 & 0.362 & \textbf{0.190} & 0.898 & 24.1 & 24.4 & 51.5 & 5721 & 2.29 & 5.83 & 1.98 & 98.0 & 82\\
\bottomrule
\end{tabular}
}
\end{table*}

To validate the efficacy and scalability of our automated annotation framework, we conducted a comprehensive evaluation across three canonical object detection benchmarks: COCO, Open Images, and Pascal VOC. These datasets were strategically selected to represent a broad spectrum of challenges, including varying object scales, annotation densities, and class distributions. The programmatic annotations generated by our framework were evaluated against the original ground truth labels, with the results detailed in Table~\ref{tab:main_results}. Our analysis compares the performance of six distinct Non-Maximal Suppression (NMS) algorithms, examining their impact on final annotation quality across key metrics, including precision, recall, F1-score, and mean Average Precision (mAP).

\subsubsection{Comparative Performance of NMS Algorithms}
The choice of NMS algorithm proves to be an essential factor in optimizing the precision-recall trade-off inherent in our high-recall, ensemble-based proposal generation system. While several methods demonstrate strong performance, nuanced differences emerge that highlight their suitability for different data characteristics.

A defining strength of our framework is its exceptionally high recall, a direct consequence of our ensemble-based weak supervision strategy. The system is intentionally prolific, generating a rich superset of candidate detections. For instance, on COCO, the pipeline generated an average of 14.20 proposals per image using standard NMS, compared to only 7.43 ground-truth objects. This results in outstanding overall recall scores (e.g., \textbf{0.802 on COCO}, \textbf{0.939 on Pascal VOC} with Soft-NMS). The primary role of the NMS algorithm is therefore to distill this high-recall, lower-precision proposal set into a high-quality final annotation set.

\textbf{Soft-NMS} consistently emerges as a top-tier performer, particularly in recall-centric metrics. It achieves the highest recall on all three datasets (COCO: \textbf{0.802}, Open Images: \textbf{0.456}, Pascal VOC: \textbf{0.939}) and secures the highest mAP across all thresholds on COCO and Open Images. Its strategy of decaying the scores of overlapping boxes rather than performing hard elimination is demonstrably superior for preserving valid detections, especially in dense scenes where objects have significant spatial overlap.

\textbf{Adaptive NMS} shows strong performance in balancing precision and recall, achieving the highest F1-score on Pascal VOC (\textbf{0.517}) and the highest precision on both COCO (\textbf{0.494}) and Open Images (\textbf{0.235}). This suggests its dynamic thresholding mechanism is effective in contexts where object densities vary.

Conversely, \textbf{Cluster NMS} consistently underperforms in F1-score while achieving the highest Average IoU across all datasets (COCO: \textbf{0.134}, Open Images: \textbf{0.128}, Pascal VOC: \textbf{0.190}). This suggests that it suppresses too many true positives—reducing recall—yet the boxes it retains are more precisely localized. Such a trade-off makes it less suitable for our goal of maximizing object discovery. In contrast, \textbf{NMS}, \textbf{DIoU NMS}, and \textbf{Weighted NMS} deliver nearly identical results across most metrics, providing reliable baselines though rarely surpassing adaptive methods like Soft-NMS.

\subsection{Performance on Challenging Data Characteristics}
A key objective of our evaluation was to assess the framework's robustness to the complexities inherent in large-scale, real-world datasets.

\vspace{1ex}
\noindent\textbf{COCO: Handling High Annotation Density and Small Objects.} The COCO dataset, characterized by its high density of objects (7.43 GT Obj/Img) and a significant proportion of small objects (48.4\% categorized as `Sm\%`), presents a formidable challenge. Our framework's ability to generate a dense proposal set (14.20 Pred Obj/Img for NMS) is crucial in this context. The success of \textbf{Soft-NMS}, which achieves the highest mAP@.75 (\textbf{0.401}) and overall mAP (\textbf{0.371}), underscoring its efficacy in resolving ambiguous overlaps in cluttered scenes without erroneously suppressing valid, closely packed objects. The high recall achieved on COCO further suggests that the multi-model ensemble is capable of identifying small objects that a single detector might miss, and the consensus mechanism effectively preserves these fine-grained detections.

\vspace{1ex}
\noindent\textbf{Open Images: Resilience to Extreme Scale and Class Imbalance.} The Open Images dataset serves as a proxy for web-scale data, with its 903 discovered classes and an extreme class imbalance ratio (up to \textbf{65,877}). In this regime, our framework's high-recall nature becomes an important discovery mechanism. While the overall F1-score is lower than on other datasets, the framework maintains a respectable recall (e.g., \textbf{0.456} with Soft-NMS). This indicates that the heterogeneous ensemble of weak labelers is effective at discovering instances even from rare, long-tail classes. However, this comes at the cost of precision, as evidenced by the high number of predicted objects per image (23.27 for Soft-NMS) compared to the number of correct detections (3.51). This highlights the fundamental challenge of managing the precision-recall trade-off in massively multi-class environments, a task where the choice of NMS algorithm is paramount. \textbf{Soft-NMS} again achieves the highest mAP, indicating a better-ranked list of detections despite the noise.

\vspace{1ex}
\noindent\textbf{Pascal VOC: Performance on High-Coverage, Simpler Scenes.} On the Pascal VOC dataset, which features fewer objects per image (2.29) and a very high ground-truth coverage rate (98.0\%), the framework demonstrates exceptional performance. The recall is outstanding, peaking at \textbf{0.939} with Soft-NMS, indicating near-complete object discovery. In this less-cluttered environment, the task of NMS is simpler. \textbf{Adaptive NMS} achieves the highest F1-score (\textbf{0.517}) by attaining the best balance between its high precision (\textbf{0.410}) and recall (\textbf{0.919}), making it the optimal choice for datasets with these characteristics.

\section{Discussion}

The \textsc{Labeling Copilot} framework demonstrates the power of an agentic approach to solve the complex, multi-stage challenge of computer vision data curation. Rather than a rigid, linear pipeline, our system is a dynamic workflow managed by a central orchestrator agent that intelligently deploys a suite of specialized tools for data discovery, synthesis, and annotation. This design philosophy is intentionally built around the core primitives of data-centric AI, ensuring the framework's longevity and adaptability.

A key architectural principle of LABELING COPILOT is the separation of fundamental data operations from their underlying model implementations. We treat Discovery, Synthesis, and Annotation as timeless primitives in the data curation lifecycle. The true innovation is not the specific models used today, but the agentic workflow that orchestrates these primitives. This separation is what makes the system robust, versatile, and future-proof. This philosophy is realized through a modular, ``hotswappable" engineering design. The agent interacts with data through standardized interfaces, such as cloud storage buckets and PASCAL VOC annotation formats. Each tool—Discovery, Synthesis, and Annotation—is a containerized module that adheres to this contract. This has profound practical implications:

\begin{itemize}
    \item \textbf{Effortless Upgrades:} The framework is not architecturally dependent on any single foundation model. For example, the current Consensus Annotation tool uses an ensemble including GroundingDINO and DETIC. Should a superior open-vocabulary detector emerge, it can be integrated simply by wrapping it in a new container that conforms to the established input/output format. The core agentic workflow requires no re-architecting.
    \item \textbf{Extensibility:} New capabilities can be added as new primitive tools. One could easily envision adding a ``Data Repair" tool that uses models to find and fix labeling errors or a ``Data Privacy" tool that automatically blurs sensitive information. The agent's capabilities can be extended without altering the existing components.
    \item \textbf{Decoupled Intelligence:} The agent’s reasoning—deciding when to synthesize more data based on the output of the discovery tool, for example—is separate from the execution of the tools themselves. This allows the agent's strategic intelligence to be improved independently of the tools' capabilities. This engineering also allows us to not overload the primary orchestrator agents' context window, allowing longer-term coherence.
\end{itemize}

The agent's ability to plan, incorporate feedback from tools, and flexibly execute these tools—sequentially, in parallel, or in iterative cycles—is the cornerstone of the framework. For example, if the Data Discovery Tool returns a sparse dataset for a particular class, the agent can autonomously decide to invoke the Synthetic Data Tool to generate new examples before proceeding to annotation. This decision-making capability allows for continuous, targeted refinement of the dataset.

This agent-driven design meets the dual requirements of generality and specificity. The agent and its core toolkit are broadly applicable to a wide range of CV tasks, from object detection to panoptic segmentation. Simultaneously, the agent's strategies and tool parameters can be customized for specific domains. By combining the outputs of its tools with human-in-the-loop feedback, the orchestrator agent learns and adapts, progressively improving the dataset in an iterative cycle. This transforms data curation from a manual, fragmented process into an intelligent, automated, and versatile solution for diverse AI challenges.
\section{Conclusion}

To address the critical bottleneck of data curation in computer vision, we introduced LABELING COPILOT, the first deep research agent for automating this complex task. Our system replaces rigid pipelines with a dynamic agentic workflow that intelligently unifies data discovery, synthesis, and annotation. Large-scale validation confirmed our approach: the Calibrated Discovery tool employs techniques that are up to 40 times more computationally efficient than optimized alternatives, and the Consensus Annotation module achieves a 37.1\% mAP on the dense COCO dataset, discovering 598 new classes and labeling twice as many objects per image. These results validate that an agentic framework built on scalable, individually optimized tools provides a robust foundation for creating industrial-scale datasets. The modular design ensures \projname{} is an extensible and versatile solution, representing a significant step forward in solving critical challenges for data-centric AI.

\bibliographystyle{IEEEtran}
\bibliography{references}

\end{document}